\documentclass[letterpaper, 10 pt, conference]{ieeeconf}  

\IEEEoverridecommandlockouts                              

\usepackage{amsmath} 
\usepackage{amssymb}  
\usepackage{bm}
\usepackage{tabularx}
\usepackage{multirow}
\usepackage{color}
\usepackage{url}
\usepackage{subcaption}
\usepackage{breqn}
\newcommand{\junk}[1]{}
\usepackage[export]{adjustbox}
\usepackage{tabulary,booktabs, multirow}
\usepackage{xcolor}
\usepackage{soul}

\usepackage{algorithmicx}
\usepackage{algorithm}
\usepackage{algpseudocode}

\usepackage[hidelinks]{hyperref}
\usepackage[noabbrev, capitalize]{cleveref}

\renewcommand{\vec}[1]{\boldsymbol{#1}}

\newcommand{\R}{\mathbb{R}}

\title{\LARGE \bf
Learning Dynamic Tasks on a Large-scale Soft Robot \\ in a Handful of Trials 

}

\author{Sicelukwanda Zwane$^{1}$, Daniel Cheney$^{2}$, Curtis C. Johnson$^{2}$, Yicheng Luo$^{1}$, Yasemin Bekiroglu$^{1,3}$,\\ Marc D. Killpack$^{2}$, Marc Peter Deisenroth$^{1}$
\thanks{$^{1}$UCL Centre for Artificial Intelligence, University College London, UK. Corresponding author email: {\tt\small sicelukwanda.zwane.20@ucl.ac.uk}}
\thanks{$^{2}$ Department of Mechanical Engineering, Brigham Young University, Provo, Utah, UT 84602, USA}%
\thanks{$^{3}$Department of Electrical Engineering, Chalmers University of Technology, Göteborg SE-41296, Sweden}%
\thanks{This work was partially supported by the National Science Foundation under Grant No. 1935312 and the Engineering and Physical Sciences Research Council (EPSRC) [EP/S021566/1]. For the purpose of Open Access, the author has applied a CC BY public copyright licence to any Author Accepted Manuscript version arising from this submission. 
The codebase associated with this paper is available at \mbox{\scriptsize \url{https://github.com/Sicelukwanda/BayesOptSoftRobotControl}}}%
}
\begin{document}

\maketitle
\thispagestyle{empty}
\pagestyle{empty}

\begin{abstract}

Soft robots offer more flexibility, compliance, and adaptability than traditional rigid robots. They are also typically lighter and cheaper to manufacture. However, their use in real-world applications is limited due to modeling challenges and difficulties in integrating effective proprioceptive sensors. Large-scale soft robots ($\approx$ two meters in length) have greater modeling complexity due to increased inertia and related effects of gravity. Common efforts to ease these modeling difficulties such as assuming simple kinematic and dynamics models also limit the general capabilities of soft robots and are not applicable in tasks requiring fast, dynamic motion like throwing and hammering. To overcome these challenges, we propose a data-efficient Bayesian optimization-based approach for learning control policies for dynamic tasks on a large-scale soft robot. Our approach optimizes the task objective function directly from commanded pressures, without requiring approximate kinematics or dynamics as an intermediate step. We demonstrate the effectiveness of our approach through both simulated and real-world experiments.

\end{abstract}

\section{Introduction}

Elephant trunks, snakes, and certain invertebrates are capable of highly dynamic and fast motion while maintaining their flexibility, prehensile nature, and compliance. However, despite being modeled after these biological entities, soft robots still struggle to perform highly dynamic control tasks. This is because soft robots are effectively infinite-dimensional dynamical systems, and it is difficult to derive accurate kinematic or dynamics models that describe their motion well enough to allow for the design of controllers that fully exploit their capabilities. 

Large-scale soft robots (a few meters in length) are attractive because of their high force-to-weight ratio \cite{li2023large}. However, their increased inertia, surface area, and related gravitational effects also present further modeling difficulties.   

\begin{figure}
    \centering
    \includegraphics[width=0.9\columnwidth]{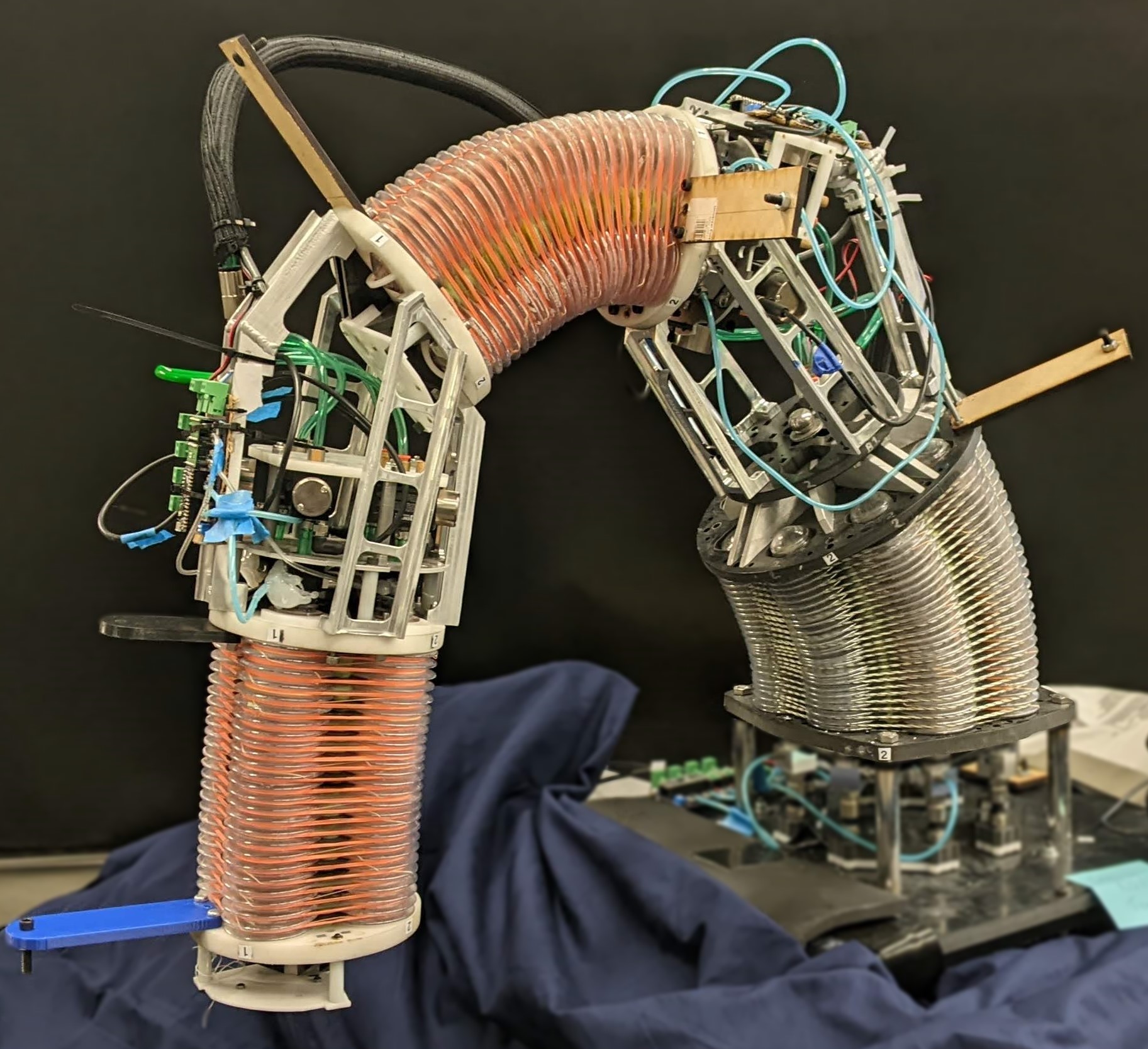}
    \caption{The soft robot design used in this paper for  simulation and hardware-based experiments. This is a 12 degree of freedom (DOF) pneumatically actuated, large-scale soft robot about 1.3 meters in length. It consists of three continuum joints connected by two rigid links.}
    \label{fig:hardware}
\end{figure}

Previous innovations in soft robot modeling, such as constant curvature \cite{hannan2003kinematics}, PDEs \cite{trivedi2008geometrically}, finite elements \cite{grazioso2019geometrically}, first-principles \cite{johnson_2021}, splines \cite{luo2020spline}, and deep learning \cite{truby2020distributed} have allowed soft robots to perform well enough at static control tasks. However, they generally require the robot to move slower to maintain lower tracking error, and fail when the robot is carrying non-uniform loads or operating at higher velocities. Furthermore, their tracking errors are higher than those of traditional rigid robots. 

Modeling approaches with high data requirements (i.e., deep neural networks), are less suitable for soft robots because the data collection process may exacerbate ``wear and tear'' (which occurs naturally and frequently over time) on the compliant parts of the robot. This may affect the robot's dynamics and cause significant deviations from expected behavior. Other effects such as temperature may also affect the robot's dynamics or contribute to measurement noise. 

Given these modeling challenges, we adopt a \emph{task-oriented}, Bayesian optimization-based control learning approach that does not require an explicit model of the soft robot. Instead, we learn a direct mapping from the controller parameters to the task objective function and use this mapping to find control parameter settings that maximize the task objective. This approach improves the efficiency of building a controller by significantly reducing the number of necessary hardware evaluations on the robot. 

The wide variety of designs and underlying actuation principles (i.e., pneumatics, hydraulics, or cable-driven mechanisms) in soft robots for different applications have also made transferring modeling and control strategies between robot platforms challenging. In this regard, the Bayesian optimization (BayesOpt) approach presented here makes fewer platform-specific assumptions, making the approach easier to adapt for use on different soft robots. 

Our contributions can be summarized as follows: 
\begin{enumerate}
    \item We present a Bayesian optimization-based approach to soft robot control tuning, capable of learning dynamic, high-dimensional behaviors directly from low-dimensional control parameterizations.
    \item We learn successful control policies with remarkably few interactions with the soft robot, minimizing possible wear and tear on the robot hardware.
    \item We evaluate the approach on two challenging tasks: throwing and hammering, which require highly dynamic motions and complex velocity profiles for successful completion.
    \item We demonstrate our approach on a physical large-scale soft robot (shown in \cref{fig:hardware}), successfully learning control parameters for a ``pseudo-throwing'' task despite uncertain objective function observations.
\end{enumerate}

\section{Hardware Description}

In this work, we consider the large-scale soft robot shown in \cref{fig:hardware}. This robot consists of three continuum joints that are connected together with rigid links. Each continuum joint is constructed with four plastic pressure chambers, which are individually pressure controlled. The chambers are arranged in an antagonistic pattern, such that each pair of pressure chambers causes a net bending torque along the longitudinal section of the robot. Each joint's bending stiffness is related to the material properties of the plastic pressure chambers. 

The chambers in each continuum joint are lengthwise constrained by an inextensible cable running through the joint's center. Externally, they are wrapped together to ensure high torsional stiffness. Although the robot is designed to resist elongation and twisting, these effects can still occur under high velocity or non-uniform loading conditions, introducing additional sources of uncertainty. 

We control the robot by issuing a vector of 12 pressure commands (4 commands for each continuum joint). The low-level controller on the robot then adjusts the pressure in each chamber to match the commanded pressures within a specified tolerance.

\section{Related Work}
In this section, we provide a brief overview of previous works on learning controllers for soft robots and discuss associated challenges in performing dynamic tasks. We categorize the approaches for soft robot control into the following areas:

\subsubsection{First-principles (FP) model-based control} 
One method to enable dynamic control is to build learned models or controllers from first-principles models. In \cite{johnson_2021}, the authors combine a learning-based method with a first-principles model to improve overall performance for a non-linear model predictive controller. However, the state space exploration approach used could be limiting and the actual efficacy for a specific task beyond trajectory tracking is not evaluated. 

\subsubsection{Data-driven model-based control} There are various approaches based on Model Predictive Control (MPC) combined with learning from data for soft robot control. For example, in \cite{hyatt_2019}, the authors combine empirical modeling (via DNNs) with model-based control, while in \cite{gillespie_2018}, the authors learn a model directly. In both cases, these learned models are used to enable model predictive control for soft robots. However, sample-efficient, gradient-based MPC tends to require a differentiable and explicit objective function. In the case of a task like hammering or throwing (as presented in this paper), this function may be difficult to define explicitly. 

\subsubsection{Model-Free Control} 
Alternatively, model-free soft robot control has been explored more thoroughly than model-based approaches \cite{Thuruthel_2017, icinco22, bianchi2023softoss}. In \cite{Thuruthel_2017}, the authors present a trajectory optimization method for predictive control of a soft robot manipulator in task space with a learned dynamic model. However, the method produces some inaccuracies in the learned model, due to the prediction of static friction effects.
In \cite{icinco22}, the authors introduce an open-loop controller for throwing tasks with a soft arm. A neural network is used to approximate the relationship between the actuation set and the target landing position. This approach has not been tested on a real robot, which would generally be more complex than in simulation. 

The approach most similar to our paper would be in \cite{bianchi2023softoss} where they use reinforcement learning to enable accurate tossing of a ball with a soft robot that starts in a default position of hanging down. However, in addition to the differences in training methods, throwing speed and distance (which exacerbate the difficulty of unmodeled, underdamped dynamics of soft robots) are not major objectives. 

Unlike aforementioned approaches, our work frames the soft robot control tuning problem as a black-box optimization problem directly in the space of pressure commands without needing an explicit dynamics model or state parameterization for the soft robot. Our Bayesian optimization-based approach efficiently explores the space of possible solutions, learning successful control policies in the smallest number of robot evaluations despite the presence of measurement noise. Moreover, the learned control policies can cope with the increased inertia of the large-scale soft robot, while also using the compliance in the joints to store and effectively release energy for throwing or hammering.

\section{Robot Control using Bayesian Optimization} 
\label{method}

In this section, we detail our approach for finding good policy parameters using Bayesian Optimization (BayesOpt). 

\subsection{Setting}
We consider finite-episodic tasks, where the environment (robot and task objects) can be ``reset'' to the same starting conditions. To avoid having to find a state description and the need for a kinematic model for the large-scale soft robot, we consider an open-loop policy $\pi(\vec \theta)$ with parameters $\vec \theta \in \R^D$. For each trial, the policy is expected to produce a sequence of controls $\vec u \in \R^{U}$ over the task horizon $H$. At the end of the trial, after executing all control commands $\vec u_{1:H} = \pi(\vec \theta)$, we receive a single observation from a task utility function 

\begin{equation}
    J( \vec \theta) = \mathbb{E}\left[ R(\vec u_{1:H}) \mid \vec \theta \right],
    \label{eqn:return}
\end{equation}
where $R(\cdot)$ may be any value indicating the performance over the entire action sequence $\vec u_{1:H}$. For example, in a throwing task, we may use a tape measure to record the final displacement of a projectile after sequentially executing individual control actions $\vec u_t$ on the soft robot. We may repeat the trial multiple times with the same parameters and average over the manual measurements to minimize measurement uncertainty.

\subsection{Problem Formulation} 
The search for optimal policy parameters $\vec \theta^*$ with an unknown objective function $J(\vec \theta)$ can be formulated as a black-box optimization problem

\begin{equation}
   \vec \theta^* = \arg \max \limits_{\vec \theta \in \R^D} J(\vec \theta).
   \label{eqn:BO}
\end{equation}

Because $J$ is unknown, we have to adopt a data-driven strategy to find $\vec \theta^*$. Moreover, this strategy has to be data-efficient since experiments are too time intensive to conduct on the physical robot, requiring hours of human time for setup, monitoring, and ``resetting'' for each objective function evaluation $J(\vec \theta)$. Furthermore, we cannot perform a large number of evaluations on the soft robot because we risk causing changes to the its dynamics. Although measuring such changes is not an easy task, we can directly observe that even the equilibrium configuration changes over the period of minutes if left in a specific configuration due to material properties. In addition, similar soft robot hardware has been shown to have hysteresis curves and pressure-to-torque output relationships that vary based on configuration and from one loading to the next (see \cite{ gillespie2016simultaneous, best2016new}). This variation of system dynamics and the need to minimize the number of hardware evaluations makes Bayesian optimization an effective approach to address these problems. 

\subsection{Bayesian Optimization based Control}
BayesOpt solves problems of the form \cref{eqn:BO} by learning a surrogate model or response surface of the objective function from data. Next, it optimizes an acquisition function $\alpha(\cdot)$ over the input space to find the next objective function input to evaluate. This input and resulting objective function observation are added to the dataset which is used to improve the surrogate model and the process is repeated.

In our robot control context, we apply this approach according to \cref{alg:BO}, with a dataset consisting of the policy parameters and corresponding objective function observations $\mathcal{D} = \{\vec \theta_i, J(\vec \theta_i)\}_{i=1}^N$. 

\begin{algorithm}
\caption{Bayesian Optimization for Robot Control}
\textbf{Input:} Task objective function $J$, domain $\R^D$, initial observations $\mathcal{D}_0 = \{(\vec \theta_i, J(\vec \theta_i))\}_{i=1}^{N_{init}}$, where $\vec \theta_i \in \R^D$ and an acquisition function $\alpha(\vec \theta)$.
\begin{algorithmic}[1]
\State Fit a GP on the initial dataset  $\mathcal{D}_0$ 
\For{$n = 1, 2, \ldots, N$}
    \State Select $\vec \theta_n$ by optimizing the acquisition function: \[\vec \theta_n = \arg\max \limits_{\vec \theta \in \R^D} \alpha(\vec \theta \mid \mathcal{D}_{n-1})\]
    \State Get action sequence from the policy  $\vec u_{1:H} = \pi(\vec \theta_n)$ \label{algo:policy_eval}
    \State Execute action sequence $\vec u_{1:H}$ on the soft robot
    \State Record task performance $J(\vec \theta_n)$ for $\vec u_{1:H}$
    \State Update the dataset $\mathcal{D}_n = \mathcal{D}_{n-1} \cup \{(\vec \theta_n, J(\vec \theta_n))\}$
    \State Update the GP model with $\mathcal{D}_n$ \label{algo:GP_update}
\EndFor
\end{algorithmic}
\textbf{Output:} The control parameters $\vec \theta^* \in \{\vec \theta_1, \ldots, \vec \theta_N\}$ with the maximum objective function value $J(\vec \theta)$.
\label{alg:BO}
\end{algorithm}

The performance of BayesOpt greatly relies on the chosen surrogate model and acquisition function. We briefly explain these concepts in the following subsections and motivate our specific choices in this work.

\subsubsection{Gaussian process Surrogate Model}

 For the dynamic, high-inertia tasks we consider in this work, we expect measurement noise in the objective function evaluations, i.e., the same set of policy parameters $\vec \theta$ may not yield exactly the same $J(\vec \theta)$. This is especially true in the case of experiments on the physical soft robot. Due to this measurement noise, the optimum of \cref{eqn:return} may not exactly coincide with the optimum of the true objective, which corresponds to the best possible task-solving behavior \cite{calandra2016bayesian}. This motivates the use of a probabilistic surrogate model, which can cope well with noisy observations. For this reason, we use a Gaussian process surrogate model.

A Gaussian process (GP) \cite{rasmussen2006gaussian} has the capacity to learn an unknown non-linear function and express it as a posterior distribution over all possible functions under a given prior distribution and training dataset. Given a set of query control parameters $\vec \theta_*$, we can obtain the mean and covariance of the GP predictive posterior in closed form as follows:

\begin{align}
    &\mu(\vec \theta_*) = \vec K_*^T \left(\vec K+\sigma_{\varepsilon}^2\vec I\right)^{-1}\vec y
    \label{eqn:GP_mean}
    \\
    &  \text{cov}(\vec \theta_*) = \vec K_{**}-\vec K^T_*\left(\vec K+\sigma_{\varepsilon}^2\vec I\right)^{-1} \vec K_*,
    \label{eqn:GP_cov}
\end{align}
where  $\vec K_{**} = k(\vec \theta_*,\vec \theta_*)$, $\vec K_{*} = k(\vec \Theta,\vec \theta_*)$, and $\vec K = k(\vec \Theta,\vec \Theta)$ are  matrices constructed using the GP kernel function $k(\cdot, \cdot)$ and training data $\vec \Theta = \begin{bmatrix} \vec{\theta}_1 & \cdots & \vec{\theta}_N\end{bmatrix}$, $\vec y = \begin{bmatrix} J(\vec{\theta}_1) & \cdots & J(\vec{\theta}_N) \end{bmatrix}$. $\vec I$ is the identity matrix, and $\sigma_{\varepsilon}^2$ is the measurement noise variance in the objective function observations $\vec y$.

 The kernel function choice is particularly important in BayesOpt because it is directly responsible for optimization-relevant properties, such as the number and frequency of local minima, differentiability (smoothness), and convexity of the learned objective function. Moreover, a naive kernel choice, which assumes the incorrect function class for $J$, may prevent the discovery of high-utility soft robot policy parameters. In this work, following the recommendations of \cite{snoek2012practical},  we use the automatic relevance determination (ARD) Matern52 kernel 
\begin{equation}
    k(\vec \theta, \vec \theta') = \sigma^2_f \left(1 + \sqrt{5d^2}+ \frac{5d^2}{3}\right) \exp \left(- \sqrt{5d^2} \right),
    \label{eqn:kern}
\end{equation}
where $\sigma_f^2$ is the signal variance and
\begin{equation*}
    d^2 = \sum_{i=1}^D\frac{(\theta_i - \theta_i')^2}{l_i^2}.
\end{equation*}

We train the GP on the latest set of observations $\mathcal{D}_n$ by maximizing the log marginal likelihood 
\begin{equation}
\begin{aligned}
    \log p(\vec{y}\mid \{\vec{\theta}_n\}^{N}_{n=1}) = &-\frac{1}{2}\vec{y}^T \left(\vec{K}+\sigma_{\varepsilon}^2\vec{I}\right)^{-1}\vec{y}\\
    &-\frac{1}{2} \log |\vec{K}+\sigma_{\varepsilon}^2\vec{I}| -\frac{N}{2}\log 2\pi
    \label{eqn:gp_nll}
\end{aligned}
\end{equation}
during the GP update step (\cref{algo:GP_update} of \cref{alg:BO}).
We use gradient-based (L-BFGS-B) multi-start optimization to optimize over the full set of GP parameters consisting of the likelihood variance $\sigma_{\varepsilon}^2$ and the Matern52 kernel hyperparameters which include the lengthscales $\vec \Lambda = [l_1, l_2, \dots, l_D ]$ and signal variance $\sigma_f^2$, training a new GP each time the dataset $\mathcal{D}$ is updated. The GP parameters are initialized according to a Gamma distribution prior, a default setting in our Botorch \cite{balandat2020botorch} implementation. In general, the performance of the GP can be expected to improve when increasing the size of the initial dataset. However, for fair comparison against other methods,  we use a minimal initial dataset size of $D+1$, where $D$ is the number of dimensions in $\vec \theta$. 

\subsubsection{Acquisition Functions} 
The acquisition function balances the trade-off between exploring input regions where there is high uncertainty about the objective function and ``exploiting'' input regions where the surrogate model predicts high objective function values. This balance is crucial to efficiently find the global optimum of the objective function with as few evaluations as possible. We consider two common acquisition functions, namely the Upper Confidence Bound (UCB)
\begin{equation}
    \alpha( \vec \theta) = \mu(\vec \theta) + \sqrt{\kappa} \sigma(\vec \theta)
    \label{eqn:UCB}
\end{equation}
and the Expected Improvement (EI)
\begin{equation}
    \alpha(\vec \theta) = \sigma(\vec \theta) \left[v\Phi(v) + \phi(v) \right]; \qquad v = \frac{F-\mu(\vec \theta)}{\sigma(\vec \theta)},
    \label{eqn:EI}
\end{equation}
where $F$ is the best objective function value so far. $\Phi(\cdot)$ and $\phi(\cdot)$ are the cumulative density function and probability density function of the normal distribution, respectively. UCB (\cref{eqn:UCB}) is the function representing some scalar (i.e., $\kappa$) number of standard deviations $\sigma(\vec \theta)$ above the mean function $\mu(\vec \theta)$ of the GP predictive posterior (\cref{eqn:GP_mean}) and $\sigma(\vec \theta)$ is the corresponding standard deviation derived from predictive posterior at $\vec \theta$ (\cref{eqn:GP_cov}). As the name suggests, EI takes an expectation of improvements (positive changes in function value) over multiple surrogate model posterior samples at a given parameter configuration $\vec \theta$, for surrogate models with Gaussian likelihoods, this simplifies to the form presented in Equation \ref{eqn:EI}. In this work, we employ the Log Expected Improvement (LEI) \cite{ament2024unexpected}, an EI variant particularly well-suited for high-dimensional problems.

\subsection{Few-parameter Control Policy} 
To perform policy evaluation (\cref{algo:policy_eval} of \cref{alg:BO}), we require an efficient mapping from policy parameters $\vec \theta \in \R^D$ to a sequence of $H$, $U$-dimensional pressure commands $\{\vec u_t \}_{t=1}^H$. One potential method is to allocate a single ``free parameter'' to each degree of freedom. For a task horizon $H=5$ on the  soft robot, this translates to 60 parameters because we need a 12 DOF pressure command at each time step in the task horizon. However, using a GP as a surrogate model imposes restrictions both on the size of the training data $\mathcal{D}$ and the dimensionality of the parameter space $\R^D$. In fact, for computational feasibility, the recommended number of dimensions in $\vec \theta$ is $D < 20$ \cite{frazier2018tutorial}. Various strategies have been presented in previous works to reduce the dimensionality of the control space when using GP-based BayesOpt.  For instance, \cite{calandra2016bayesian} discretized the continuous parameters of a bipedal robot's leg joints into three states—\textit{flex}, \textit{hold}, and \textit{extend}—before fine-tuning torque and timing for switching over these states using Bayesian Optimization. Similarly, \cite{cully2015robots} confined the actuation of each leg of a spider robot to five discrete settings. 

Adopting a parallel strategy, we discretize our soft robot's command pressures into $P$ distinct values spanning from minimum to maximum pressures. Under the discretization $P$, we generate the set of all possible actions $\mathcal{A}_P = \{\vec u_i\}_{i\in \mathcal{I}}$ with $\mathcal{I}\subset \mathbb{N}$ serving as the index set. Our control policy, $\pi \circ g$, is defined in this indexed space, mapping parameters $\vec{\theta} \in \mathbb{R}^D$ to an action sequence within $\mathcal{A}_P$. Here, $g: \vec{\theta} \to \mathcal{I}^H$ is a transformation that converts the continuous parameters into $H$ indices from the set $\mathcal{I}$. Unlike the ``free parameter'' setting, our index set control policy only requires $D=H$ parameters. 

The problem of choosing a sequence of $H$ action indices from a set of $M = |\mathcal{A}_P|$ actions has $M^H$ possible solutions (i.e., action sequences). We condense the size of this search space by pairing pressure chambers such that we let $\rho_i = \rho_{\text{max}} - \rho_{j}$ for a given antagonistic pair $(\rho_i, \rho_{j}) \in \vec u$ on our large scale soft robot (See \cref{fig:hardware}). This reduces the overall degrees of freedom from $U=12$ to $U=6$. However, evaluating all action sequences on the physical soft robot to find the most performant one is still impractical even for small values of $P$. For example, $P=2$ and $H=5$ results in $M = 64$ actions and approximately $10^9$ possible action sequences $\vec u_{1:H}$, corresponding to about 6 years of non-stop robot time for a command rate of $5$ Hz. 
As such, we employ BayesOpt to efficiently search for ``objective-maximizing'' policy parameters in a handful of trials.   

\section{Simulation Experiments and Results}
\label{experiments}

\subsection{Simulation Description}

\begin{figure}
\begin{center}
\centerline{\includegraphics[height=4.5cm]{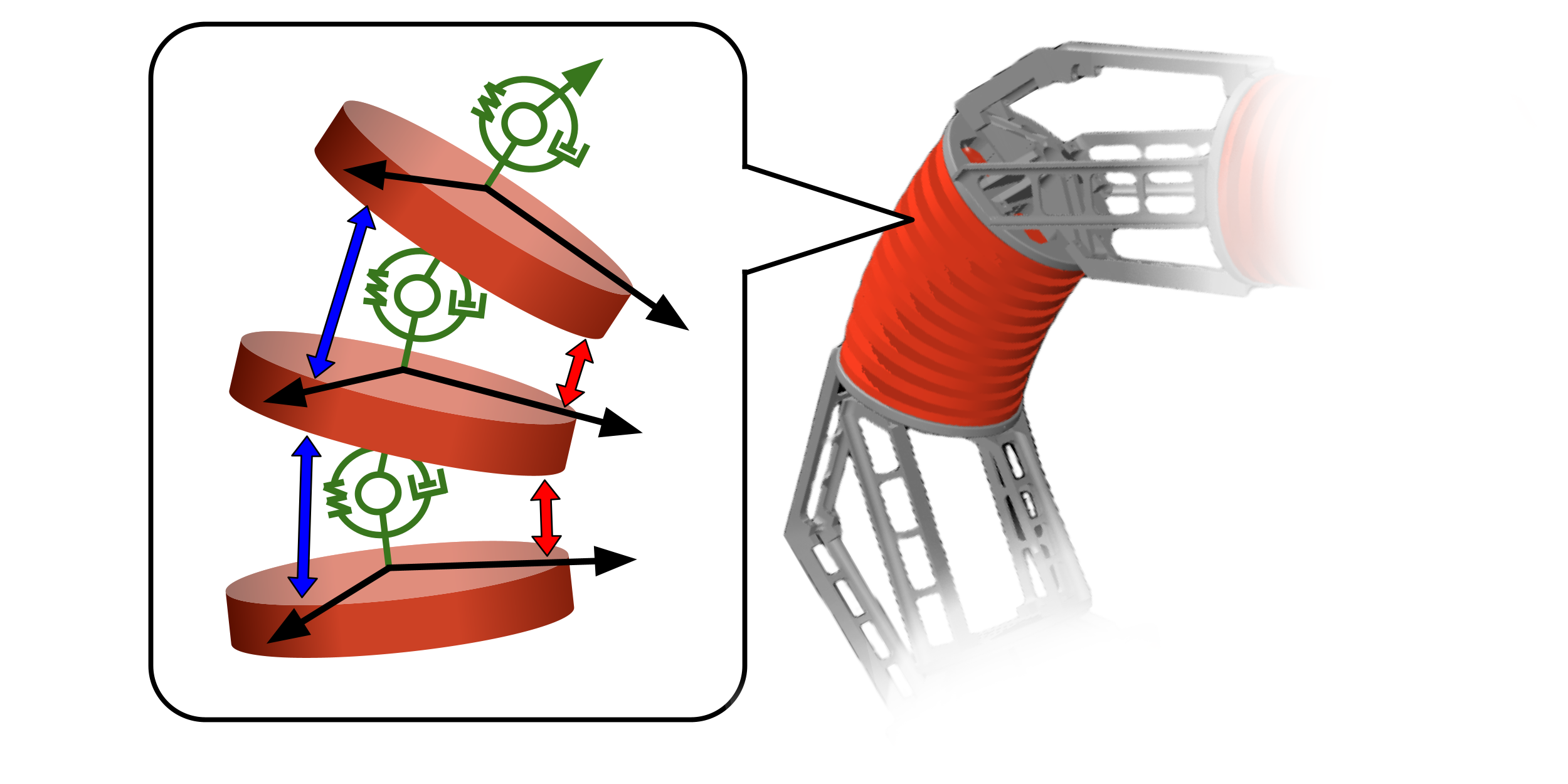}}
\caption{An illustration of the simulated version of the physical robot. Each continuum joint is approximated by individual disks that are connected by universal joints with springs and dampers connected in parallel, shown in green. Compressive forces from pneumatic actuators are applied between disks, along the $x$ and $y$ axes, shown in red and blue, respectively.}
\label{fig:mujoco-sim}
\end{center}
\end{figure}

We use MuJoCo \cite{todorov2012mujoco} to simulate the physical soft robot for the simulation experiments. We approximate each continuum joint with a series of thin disks, as shown in \cref{fig:mujoco-sim}. The disks are kinematically linked using universal joints and are driven by simulated pneumatic chambers which apply compressive forces (red and blue arrows) between the disks to cause bending about the $x$ and $y$ axes of the joint. For clarity in \cref{fig:mujoco-sim}, we show only two compressive forces in between each disk but there are four forces in reality: two pairs of antagonistic forces, one red pair and one blue pair. The forces from each pair result in a net torque that causes bending about one of the axes of the universal joint. We model the compliance of the pneumatic chambers with a spring and damper in parallel with each universal joint. The spring acts as a parasitic stiffness and attempts to pull the continuum joint back to a nominal curvature.
We use the simulator to evaluate the performance of our BayesOpt pipeline, run ablation studies, and compare different policy optimization approaches. Additionally, the simulator allows for the visualization of the policy. This gives the human operator an idea of what kind of motion to expect when carrying out experiments on the physical robot and prepare safety measures in advance.

\subsection{Dynamic Task Descriptions}

Below, we detail dynamic tasks that leverage the flexible characteristics of the soft robot, but also showcase the difficulty of finding effective mappings between control inputs and effective performance.
\subsubsection{Throwing Task}
For this task, we attach a prismatic gripper to the simulated robot, allowing it to hold and release a $0.25$ kg ``life-size'' object. We include an additional parameter $t_{\text{R}} \in \left[ 1, \dots, H\right]$, which corresponds to the release time of the cube and set up an objective function 
\begin{equation}
    J(\vec \theta)= \sum_{t=1}^{t_{\text{R}}}
            \lVert \dot{\vec x}_t \rVert  + \vec 1_{H}(t)\cdot D_{\text{cube}} 
    \label{eqn:rew-throw}, 
\end{equation}
where $\dot{\vec x_t}$ corresponds to the end-effector velocity and $D_{\text{cube}}$ is the displacement in meters of the thrown object from the base of the robot. $\vec 1_{H}(t)$ is an indicator function which returns $1$ when $t=H$ and zero elsewhere.

\begin{figure}
    \centering 
    \vspace{0.5cm}
    \begin{subfigure}[b]{0.47\columnwidth}
        \includegraphics[height=3.7cm, trim=15pt 20pt 0pt 50pt]{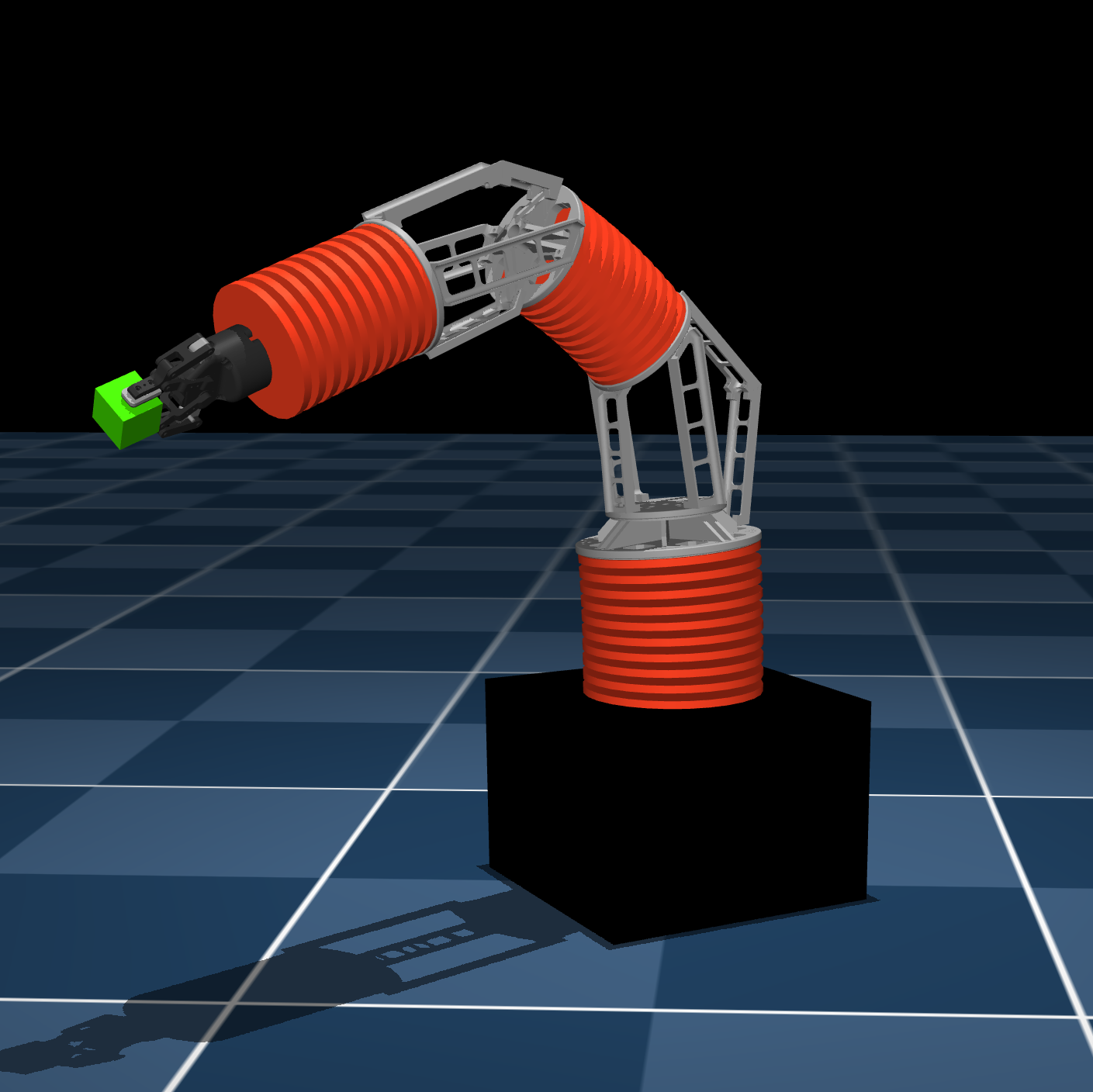}
        \caption{}
        \label{fig:sim-throw:a}
    \end{subfigure}
    \begin{subfigure}[b]{0.47\columnwidth}
        \includegraphics[height=3.7cm, trim=15pt 20pt 0pt 50pt]{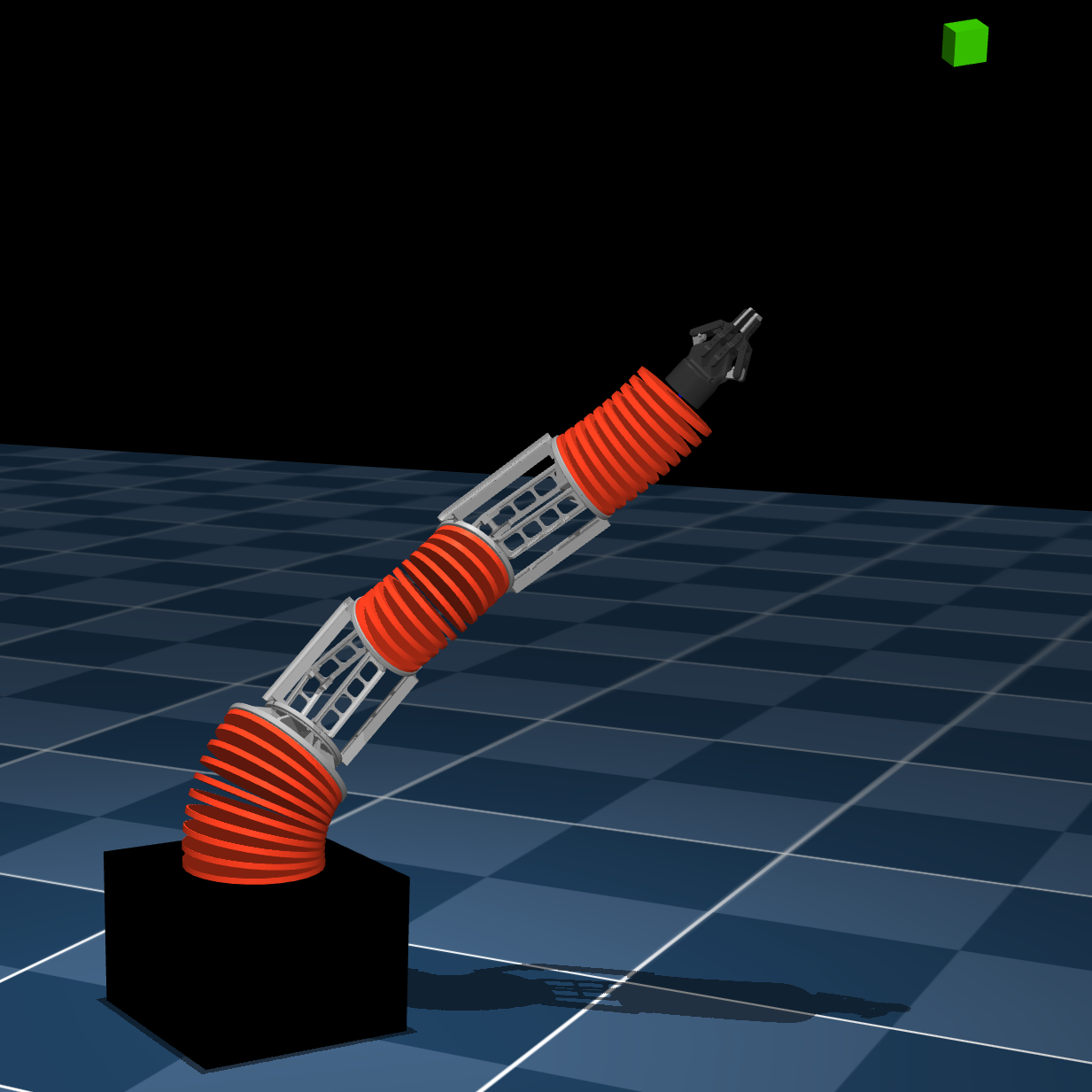}
        \caption{}
        \label{fig:sim-throw:b}
    \end{subfigure}
    \caption{Throwing task: The robot starts with the gripper holding the object \ref{fig:sim-throw:a}), generates high end-effector velocity, and has to release the object the right time \ref{fig:sim-throw:b}) to throw successfully.} 
    \label{fig:sim-throw} 
\end{figure}

We chose to throw a cube to minimize any rolling effects. The soft robot starts each trial with the object attached to the end-effector (\cref{fig:sim-throw:a}), performs the prescribed pressure commands, and releases the object when time step $t=t_{\text{R}}$ is reached (\cref{fig:sim-throw:b}). We use BayesOpt to find policy parameters $\vec \theta = [\theta_1, \dots, \theta_H, t_{\text{R}}]$ that maximize \cref{eqn:rew-throw}.

\begin{figure}
    \centering 
    \begin{subfigure}[b]{0.47\columnwidth}
        \includegraphics[height=3.7cm, trim=15pt 20pt 0pt 50pt]{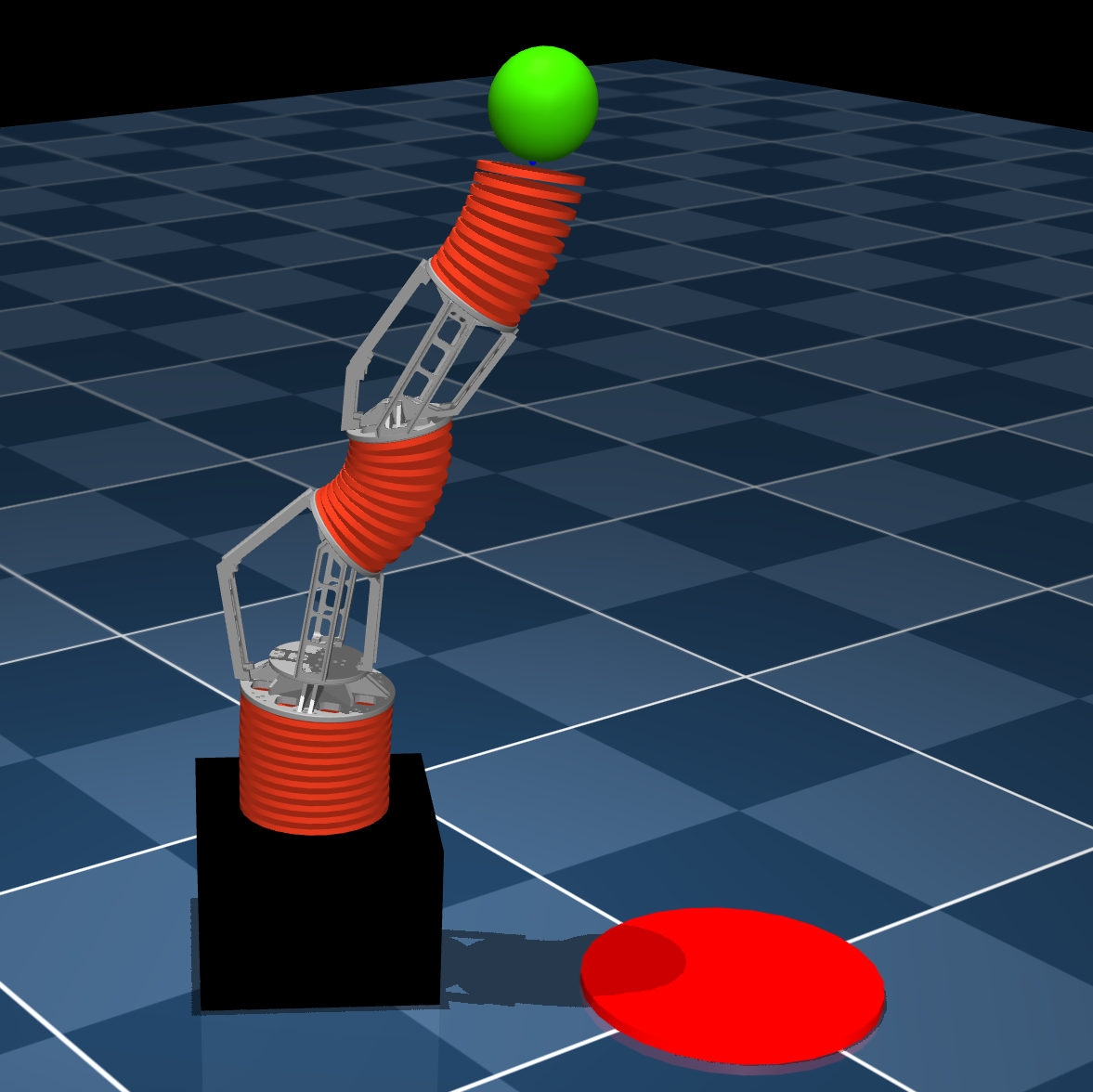}
        \caption{}
        \label{fig:sim-hammer:a}
    \end{subfigure}
    \begin{subfigure}[b]{0.47\columnwidth}
    \includegraphics[height=3.7cm, trim=15pt 20pt 0pt 46pt]{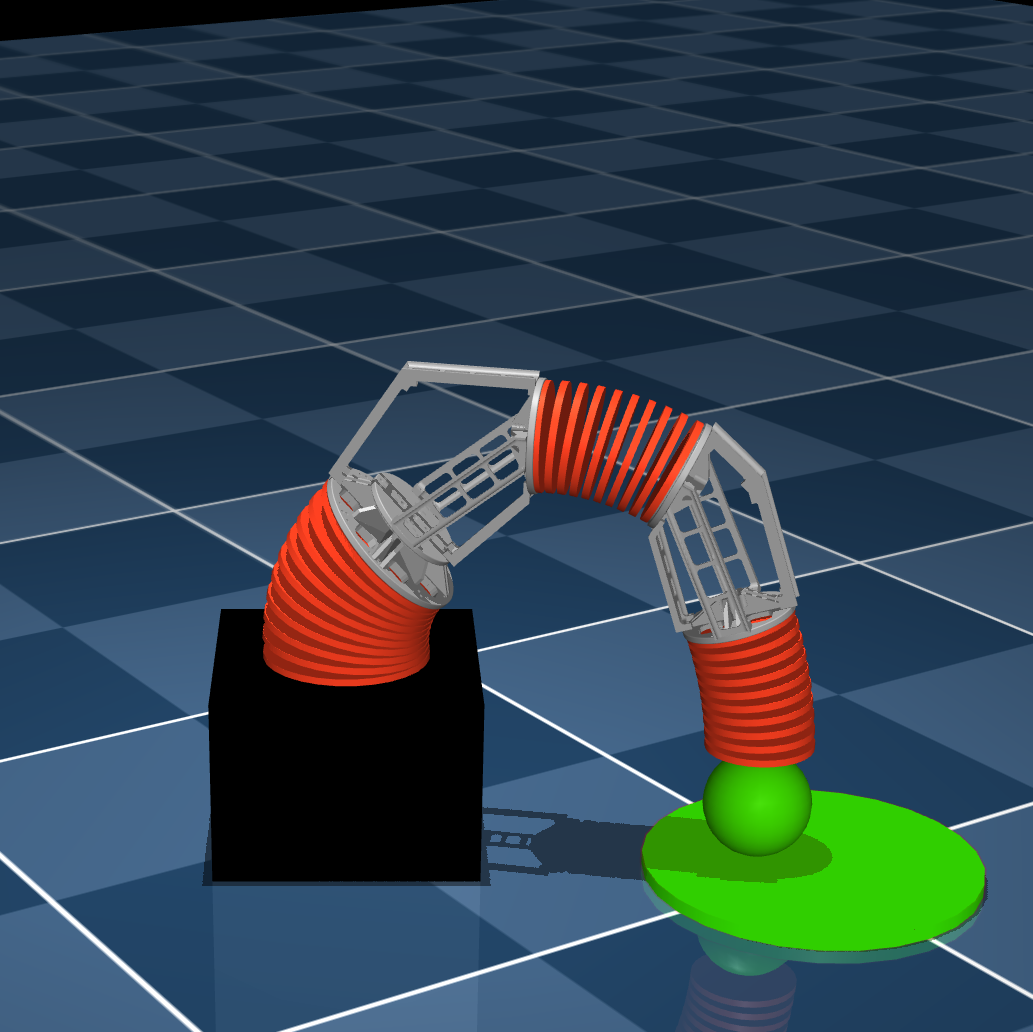}
        \caption{}
        \label{fig:sim-hammer:b}
    \end{subfigure}
    \caption{Hammering task. Starting from the equilibrium state (\cref{fig:hardware}), the robot has to ``wind-up'' enough energy \ref{fig:sim-hammer:a}) to activate the touch force sensor \ref{fig:sim-hammer:b}).} 
    \label{fig:sim-hammer} 
    
\end{figure}
\subsubsection{Hammering task}
To set up a hammering task, we attach a 2kg mass to the tip of the robot to use as the ``hammer'' and place a force sensor (diameter = 0.5 m) near the simulated soft robot at a fixed Cartesian position $\vec g = [-0.5, 0.5, 0.0]$. To activate the sensor as shown in \cref{fig:sim-hammer:b}, the robot has to exert a minimum force of 5N. For this task, the objective function 
\begin{equation}
    J(\vec \theta) =  \sum_{t=1}^H  m\vec \dot{\vec x}^{z}_t + F_t - \vec 1_{t_0:H}(t)\cdot\lVert \vec x_t - \vec g \rVert_2 
    \label{eqn:rew-hammer}
\end{equation}
is a cumulative sum of the vertical momentum of the hammer $m\vec \dot{\vec x}^{z}_t$ and the measured impact force $F_t$ for $t < t_0$. When $t\geq t_0$, we add a Euclidean distance between the hammer and the force sensor, encouraging the robot to make contact with the sensor. We found $t_0 = 0.7H$ to work well in our experiments. 

The added convenience of using a simulator means that all state information, such as the position of the hammer $\vec x_t$, can be obtained directly without state estimation or kinematics computation. 

\subsection{Experiment Setup} 
In this section, we introduce alternative global optimization strategies to compare with our BayesOpt approach: the Cross Entropy Method (CEM) \cite{botev2013cross} and Random Search. These methods are configured as follows:

\subsubsection{CEM} 
We set the population size to 50 trials and the elite percentage to 20\%. To determine the initial population distribution parameters, we randomly sample $D+1$ trials, compute a sample mean and artificially set the variance to cover all possibilities for $\vec \theta$ in $\R^D$.
\subsubsection{Random Search}
Similar to CEM, Random Search searches through a population of 50 trials at each optimization step. Unlike BayesOpt which has access to a surrogate model for the objective, both CEM and Random Search have to evaluate all candidates in the population on the robot.  

We focus on the pressure command discretization hyperparameter $P$. In general, higher $P$ allows for more granular pressure commands and may result in the discovery of better control sequences $\vec u_{1:H}$. However, the size of the policy parameter search space $M^H$ (where $M = |\mathcal{A}_P |$) also increases greatly.
For all experiments, we set a task horizon of $H=10$ over a maximum time of 5 seconds. The transformation $g$ scales the parameter values to the range $[0,M]$, and ``floors" to the nearest integer, yielding an index from the set $\mathcal{I}$.

\subsection{Results}

\begin{table*}[ht]
\vspace{2mm}
\centering
\begin{tabular}{@{}lc|cccc@{}}
\toprule
                                                                             & \# Trials & $P=2$              & $P=5$                & $P=7$               & $P=10$             \\ \hline \hline
\multirow{2}{*}{BayesOpt-UCB ($\kappa = 0.9$)}                                            & 100       & $28.526 \pm 4.025$ & $25.959 \pm 2.86$    & $21.842 \pm 7.434$  & $21.235 \pm 6.135$ \\
                                                                                    & 500       & $\mathbf{34.176 \pm 5.417}$ & $29.089 \pm 3.664$   & $26.016 \pm 7.86$   & $28.434 \pm 5.998$ \\ \hline
\multirow{2}{*}{BayesOpt-LEI}                                                             & 100       & $30.369 \pm 6.864$ & $\mathbf{30.345 \pm 6.554}$   & $\mathbf{32.208 \pm 3.709}$  & $\mathbf{31.506 \pm 4.541}$ \\
                                                                                    & 500       & $\mathbf{37.697 \pm 1.679}$ & $\mathbf{38.285 \pm 2.581}$   & $\mathbf{40.878 \pm 1.375}$  & $\mathbf{40.171 \pm 1.834}$ \\ \hline
\multirow{3}{*}{CEM}                                                                & 100       & $28.937 \pm 2.758$ & $20.730 \pm 0.918$   & $19.724 \pm 1.987$  & $20.629 \pm 2.061$ \\
                                                                                    & 500       & $33.653 \pm 1.788$ & $24.03  \pm 2.626$   & $22.731 \pm 2.793$  & $22.867 \pm 3.424$ \\
                                                                                    & 10K       & $\mathbf{41.942 \pm 41.9}$  & $\mathbf{38.931 \pm 1.897}$   & $\mathbf{36.939 \pm 2.756}$  & $\mathbf{35.412 \pm 2.407}$ \\ \hline
\multirow{3}{*}{Random Search}                                                      & 100       & $27.435 \pm 1.349$ & $22.125 \pm 1.763$   & $19.167 \pm 1.795$  & $20.614 \pm 3.245$ \\
                                                                                    & 500       & $30.354 \pm 1.835$ & $23.840 \pm 1.693$   & $22.325 \pm 1.074$  & $22.269 \pm 1.921$ \\
                                                                                    & 10K       & $34.131 \pm 1.473$ & $28.834 \pm 1.456$   & $26.027 \pm 0.493$  & $26.634 \pm 2.962$ \\ \bottomrule
\end{tabular}

\caption{Simulated throwing performance --- Best solution so far, averaged across 5 random seeds. The top three methods for each discretization setting $P$, are marked in bold.}
\label{exp:abl-throw}
\end{table*}
\begin{table*}[ht]
\centering
\begin{tabular}{@{}lc|cccc@{}}
\toprule
                                                                             & \# Trials & $P=2$              & $P=5$              & $P=7$              & $P=10$             \\ \hline \hline
\multirow{2}{*}{BayesOpt-UCB ($\kappa = 0.9$)}                                            & 100       & $1.137 \pm 0.476$  & $1.342 \pm 0.624$  & $2.588 \pm 1.019$  & $1.601 \pm 1.294$  \\
                                                                                    & 500       & $1.775 \pm 0.403$  & $\mathbf{2.552 \pm 0.7077}$ & $\mathbf{2.719 \pm 1.06}$   & $2.278 \pm 0.9145$ \\ \hline
\multirow{2}{*}{BayesOpt-LEI}                                                             & 100       & $\mathbf{2.566 \pm 0.666}$  & $1.923 \pm 0.938$  & $\mathbf{2.662 \pm 0.964}$  & $\mathbf{2.839 \pm 0.833}$  \\
                                                                                    & 500       & $\mathbf{3.387 \pm 0.3702}$ & $\mathbf{4.217 \pm 1.097}$  & $\mathbf{3.534 \pm 0.1906}$ & $\mathbf{3.541 \pm 0.42}$   \\ \hline
\multirow{3}{*}{CEM}                                                                & 100       & $1.182 \pm 0.473$  & $1.140 \pm 0.151$  & $1.157 \pm 0.184$  & $1.164 \pm 0.319$  \\
                                                                                    & 500       & $1.493 \pm 0.271$  & $1.378 \pm 0.268$  & $1.303 \pm 0.287$  & $1.232 \pm 0.318$  \\
                                                                                    & 10K       & $2.174 \pm 0.798$  & $2.026 \pm 0.603$  & $2.342 \pm 0.635$  & $2.171 \pm 0.536$  \\ \hline
\multirow{3}{*}{Random Search}                                                      & 100       & $1.441 \pm 0.621$  & $1.137 \pm 0.135$  & $1.150 \pm 0.073$  & $1.076 \pm 0.118$  \\
                                                                                    & 500       & $2.030 \pm 0.215$  & $1.438 \pm 0.345$  & $1.208 \pm 0.164$  & $1.236 \pm 0.131$  \\
                                                                                    & 10K       & $\mathbf{3.002 \pm 0.541}$  & $\mathbf{2.441 \pm 0.233}$  & $2.369 \pm 0.082$  & $\mathbf{2.281 \pm 0.303}$  \\ \bottomrule            
\end{tabular}
\caption{Simulated hammering performance --- Best solution so far, averaged across 5 random seeds. The top three methods for each discretization setting $P$, are marked in bold.}
\label{exp:abl-hammer}
\end{table*}
In \cref{exp:abl-throw} and \cref{exp:abl-hammer}, we present performance and sample-efficiency comparisons between BayesOpt, CEM, and Random Search. We observe that BayesOpt-UCB and BayesOpt-LEI perform comparably or surpass CEM and Random Search, particularly for limited-data scenarios (100 and 500 trials). For $P=2$, all methods yield similar performance after 100 trials in the throwing task. This is possibly due to there being many sub-optimal solutions. However, the hammering task is less forgiving and mainly rewards contact between the robot and the force sensor. For this task, BayesOpt with LEI is superior.
As size of the search space, $P$ increases, we see the benefits of the LEI acquisition function of being able to efficiently explore in high dimensions as highlighted in \cite{ament2024unexpected}. This suggests that LEI is well-suited for optimizing high-dimensional control problems in this domain. We found that UCB acquisition (with $\kappa = 0.9$) function also performs reasonably well across all settings of $P$ for both tasks, but there were settings where smaller values of $\kappa$ worked better. The higher value for $\kappa$ improved exploration but also increased the standard deviation in performance. A video showing the final policies for throwing and hammering in simulation can be seen at \url{https://youtu.be/OHcxJl7rC4E}. 

\begin{table*}[ht]
\centering
\begin{tabular}{cc|cc|cc}
\toprule
                 & \multirow{2}{*}{\# Trials} & \multicolumn{2}{c|}{Without Approximate Joint States}       &  \multicolumn{2}{c}{With Approximate Joint States}  \\[0.5pt]  
\cline{3-6} 
                 & & Throw              & Hammer            &  Throw              & Hammer            \\[0.5ex] \hline \hline 
\multirow{3}{*}{REDQ} & 250       & $14.782 \pm 2.944$ & $0.116 \pm 0.07$ & $15.719 \pm 2.771$ & $0.239 \pm 0.209$ \\
                      & 500       & $27.362 \pm 0.862$ & $1.411 \pm 0.125$ & $30.35 \pm 1.812$ & $2.098 \pm 0.421$ \\
                      & 10K       & $33.016 \pm 0.975$ & $2.344 \pm 0.06$ & $37.72 \pm 2.223$ & $3.12 \pm 0.065$ \\ \hline
\end{tabular}
\caption{Simulated results for the REDQ agent with and without the presence of approximate joint state information --- Best solution so far, averaged across 5 random seeds.}
\label{exp:redq_uv}
\end{table*}

\subsection{Reinforcement Learning Experiments}
The absence of rich state information makes applying traditional feedback control approaches, including reinforcement learning (RL), particularly challenging in this setting.  Nonetheless, we investigated the use of the recent REDQ \cite{chen2021randomized} agent, a data-efficient, model-free RL algorithm.

For the REDQ agent, we provided end-effector velocities, positions of task-relevant objects, and step-wise rewards derived from the objective function. Both the policy and critic use a two-layer neural network with a hidden size of 256. We use an ensemble size of 2 for the critic.
We use a high Update-To-Data (UTD) ratio of 16, and, for fair comparison, provided 2500 transitions (250 trials) as initial pre-training data. Using the same settings, we also investigated the impact of adding information about the arm's configuration (i.e., a proxy for stored energy) on performance. This information can be easily obtained in the ``ideal'' simulated environment but would be too simplistic for the real robot.

Our findings (\cref{exp:redq_uv}) indicate that REDQ struggles to learn effective policies for both dynamic tasks when given a limited data budget and restricted state information.

The results for simulated experiments show that BayesOpt is a viable strategy for controlling soft robots in dynamic task scenarios where the robot dynamics are complex. Its sample efficiency minimizes the required number of physical trials, preserving robot integrity and making BayesOpt highly attractive for real-world soft robot applications.

\section{Real-robot Experiments}

\subsection{Hardware Setup}
In the hardware experiments, task-relevant information such as the tip position and velocity was measured directly by an HTC Vive Tracker that was attached via a blue 3D printed fixture (shown in Figure \ref{fig:hardware}). Because of the bending limits on each joint and the use of a Vive Tracker sensor that could be damaged from high velocity impacts, we selected a nominal pressure value for all twelve pressure commands. These nominal pressures caused the tip of the robot to begin in a position that was lifted away from the base of the robot (although we did not control the robot to an exact joint position). Any commands that were then generated by BayesOpt or randomized policies used for initialization were added to these nominal pressures. 

In addition, for the safety of the physical robot platform, we only allowed BayesOpt to vary the pressure in maximum steps of 30 kPa (despite the maximum source pressure available being closer to 205 kPa). In terms of potential velocity imparted to an object, this constraint is fairly limiting. However, we were still able to achieve velocities over 4 m/s at the tip of the robot.

To simplify the setup and experimental procedure on the hardware, we did not attach an object or gripper that would release the object at a given time. Instead, our objective function was simply the maximum magnitude of velocity measured across the five second trajectories. Although this is not equivalent to throwing, the velocity at the tip should be highly correlated with the velocity imparted to an object when thrown. It therefore allows us to validate whether or not BayesOpt can find an effective policy on hardware, despite the significant nonlinear dynamics and complicated mapping between the input and reward space. 

Because of the complexity of running randomized and high-speed trajectories on hardware, we only allowed BayesOpt to execute 12 random policies for each acquisition function (LEI and UCB), before then executing 60 iterations in each case. Because of some electrical issues (related to I2C communication) caused by moving at high velocities, we could not collect statistics associated with multiple trials. 

\subsection{Hardware Experiment Results}
To successfully maximize the robot tip velocity, BayesOpt has to find pressure commands within $\mathcal{A}_P$ that are temporally correlated, (i.e. higher velocities are only possible by storing energy and then releasing it over the five second window). 

\begin{figure}[ht]
\begin{center}
\includegraphics[width=0.98\linewidth, trim={1.1cm, 0, 1.2cm, 0}, clip]{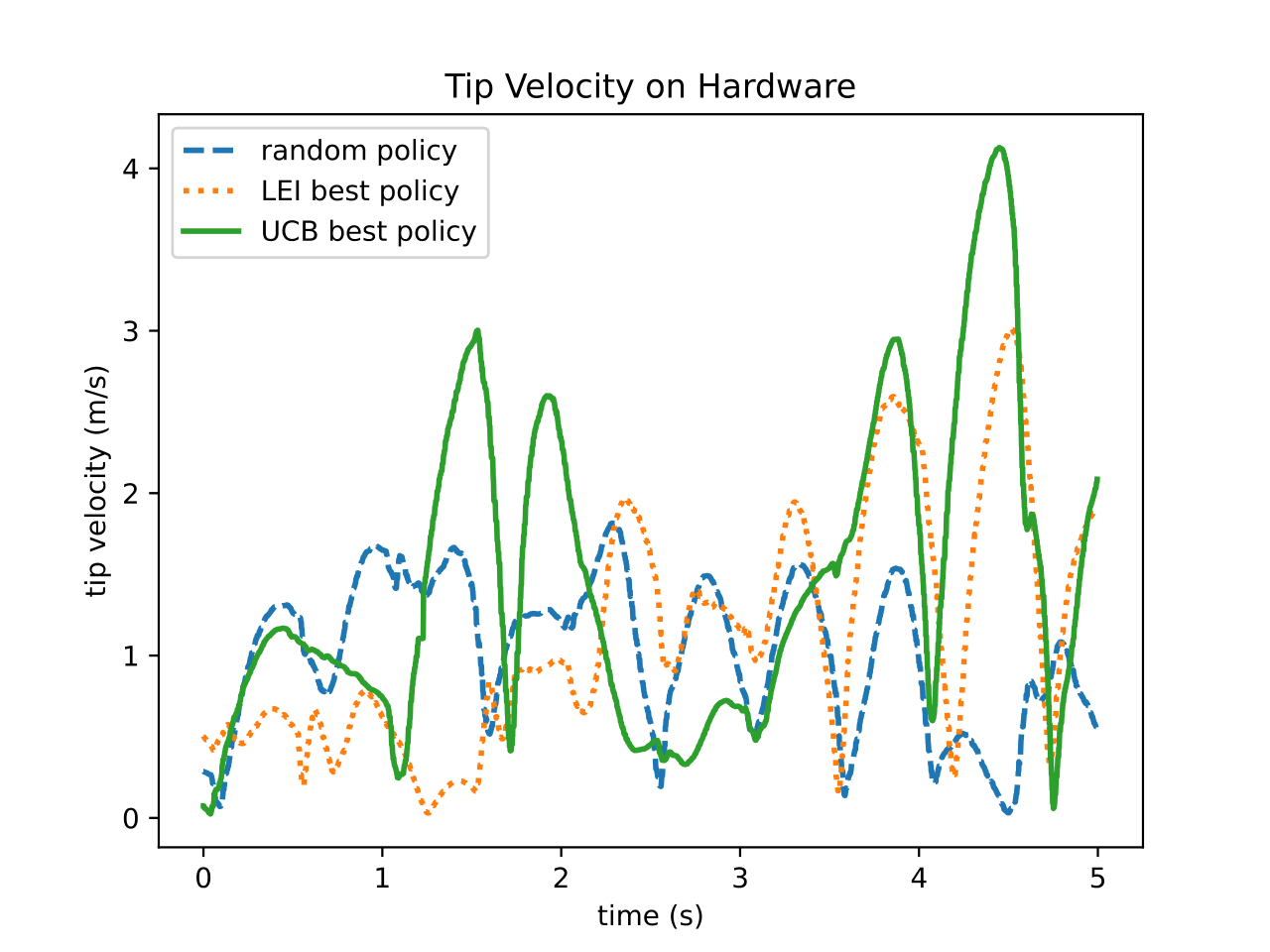}
\caption{Tip velocities that resulted from using BayesOpt-LEI,  BayesOpt-UCB, and a random policy.}
\label{exp:real-results}
\end{center}
\end{figure}

Unlike in simulation, the UCB acquisition function produced a better overall policy than the LEI acquisition function. However, both significantly increased the tip velocity relative to random policies. In \cref{exp:real-results}, we show the tip velocity over time for a random policy (used in initialization for BayesOpt), the best LEI policy, and the best UCB policy. As presented in that figure, the best policy for UCB produced a tip velocity of over 4 m/s. A video showing the entire optimization sequence for the UCB acquisition function trial can be seen at \url{https://youtu.be/OHcxJl7rC4E}.

\section{Discussion}
Like any optimizer given direct access to the pressure commands sent to the robot, the Bayesian optimization based controller can discover the highest possible tip velocities allowable by the robot hardware. However, operating the robot at high speeds and performing dynamic tasks poses safety risks. With BayesOpt, safety constraint terms can be incorporated directly into the objective, shaping the learned policies to prioritize both performance and safety. Recent advances in constrained and safely exploring Bayesian optimization \cite{ament2024unexpected, berkenkamp2023bayesian}, which are specialized for safety-critical applications, promise better results with minimal constraint violations. 
The BayesOpt control learning presented in \cref{alg:BO} searches for parameters that are a continuous relaxation over the discrete search space $\mathcal{I}$. Although this method is effective for the dynamic tasks we consider, alternative BayesOpt approaches for mixed search spaces \cite{ru2020bayesian} or for defining kernels in discrete spaces \cite{oh2019combinatorial} also exist. We leave investigations regarding safety constraints and direct discrete space optimization to future work.

\section{Conclusion}
In addition to increased load-bearing capacity, large-scale soft robots also offer inherent advantages in terms of flexibility, compliance, and impact-resistance for challenging tasks in complex environments. Although these properties make them capable of dynamic maneuvers, inherent modeling difficulties present hurdles for real-world deployment.  In this work, we have demonstrated a Bayesian optimization (BayesOpt) approach to efficiently learn dynamic, high-dimensional behaviors from simple, low-dimensional controller parameterizations for a large-scale soft robot arm. We bypass explicit kinematic and dynamic modeling, and instead focus directly on optimizing task performance.

We compare against other control policy parameter tuning approaches, which yield good performance when plenty of robot data is available. However, data-efficiency is a high priority in soft robots as the dynamics can exhibit variations and uncertainties over extended use. Our results show that BayesOpt excels in this low-data regime, where precise robot dynamics models are unavailable. Moreover, the simulation results translate well onto the physical soft robot, demonstrating that BayesOpt is able to maximize the tip velocity with only a handful of trials as training data.

\bibliographystyle{IEEEtran}
\bibliography{main}


\end{document}